\pdfoutput=1

\documentclass[11pt]{article}

\usepackage[final]{acl}

\usepackage{times}
\usepackage{latexsym}

\usepackage{amsmath}
\usepackage{multirow}
\usepackage{booktabs}
\usepackage{subcaption}
\usepackage[T1]{fontenc}

\usepackage[utf8]{inputenc}

\usepackage{microtype}

\usepackage{inconsolata}

\usepackage{graphicx}
\usepackage{tabularx}
\usepackage{xcolor}
\usepackage{soul}
\usepackage{float}
\usepackage{algorithm}
\usepackage{algorithmic}
\usepackage[normalem]{ulem}

\renewcommand{\thefootnote}{\ifcase\value{footnote}\or *\or $\dagger$\or $\ddagger$\fi}

\newcommand{\internship}{\footnotemark[1]}      
\newcommand{\equalfirst}{\footnotemark[2]}     
\newcommand{\corresponding}{\footnotemark[3]}   

\title{Visual Self-Refinement for Autoregressive Models}

\author{Jiamian Wang$^{1}$\internship\equalfirst,
Ziqi Zhou$^{1}$\equalfirst,
Chaithanya Kumar Mummadi$^{2}$,\\
{\bf Sohail Dianat$^{1}$},
{\bf Majid Rabbani$^{1}$},
{\bf Raghuveer Rao$^{3}$},
{\bf Chen Qiu$^{2}$}\corresponding
\and
{\bf Zhiqiang Tao$^{1}$}\corresponding\\
$^{1}$Rochester Institute of Technology, 
$^{2}$Bosch Research,
$^{3}$DEVCOM Army Research Laboratory
}

\begin{document}
\maketitle

\footnotetext[1]{Work partially done during an internship at Bosch.}
\footnotetext[2]{Equal contributions.}
\footnotetext[3]{Corresponding authors: Chen Qiu and Zhiqiang Tao}

\begin{abstract}
Autoregressive models excel in sequential modeling and have proven to be effective for vision-language data. However, the spatial nature of visual signals conflicts with the sequential dependencies of next-token prediction, leading to suboptimal results. This work proposes a plug-and-play refinement module to enhance the complex spatial correspondence modeling within the generated visual sequence. This module operates as a post-pretraining step to jointly refine all generated tokens of autoregressive model, enhancing vision-language modeling under a shared sequential prediction framework. By leveraging global context and relationship across the tokens, our method  mitigates the error accumulation issue within the sequential generation. Experiments demonstrate that the proposed method improves the generation quality, enhancing the model's ability to produce semantically consistent results.
\end{abstract}
\section{Introduction}\label{sec: introduction}

Autoregressive (AR) models have achieved remarkable success in recent years across multimodal tasks~\cite{brown2020language,achiam2023gpt,touvron2023llama,team2023gemini,qwen,bai2024sequential,yu2024randomized,sun2024autoregressive,el2024scalable,Sun_2025_ICCV}. Large language models (LLMs) based on autoregressive modeling~\cite{brown2020language,achiam2023gpt,touvron2023llama,team2023gemini} encode text as sequences of tokens and predict each token sequentially based on the preceding tokens. This next-token prediction paradigm effectively captures sequential dependencies and complex semantic relationships in the text, and shown to excel in tasks like question answering~\cite{trischler2016newsqa,minaee2021deep,sun2024stllava,sun2024sq} and text generation~\cite{hendryckstest2021,hendrycks2021ethics}.

Building on this success,  recent works have extended 
autoregressive modeling to visual data~\cite{bai2024sequential,yu2024randomized,tian2024visual,sun2024autoregressive,el2024scalable}. Among them, Large Vision Model (LVM)~\cite{bai2024sequential}
stands out for its focus on in-context tasks. LVM encodes images (or video frames) as a sequence of tokens and uses next-token prediction to solve various image translation and generation tasks. 
The focus on in-context learning makes LVM well-suited to handle vision tasks that require contextual information for generating coherent and consistent outputs.

Unlike text, visual tokens exhibit strong spatial correlations rather than purely sequential dependencies. This spatial structure necessitates attention to both prefix and future tokens for coherent generation. Consequently, generating tokens solely based on preceding tokens may compromise global contextual awareness, and the sequential nature of generation further exacerbates this by allowing early errors to  accumulate, degrading both visual fidelity and semantic consistency.

\begin{figure*}[t]
  \includegraphics[width=\linewidth]{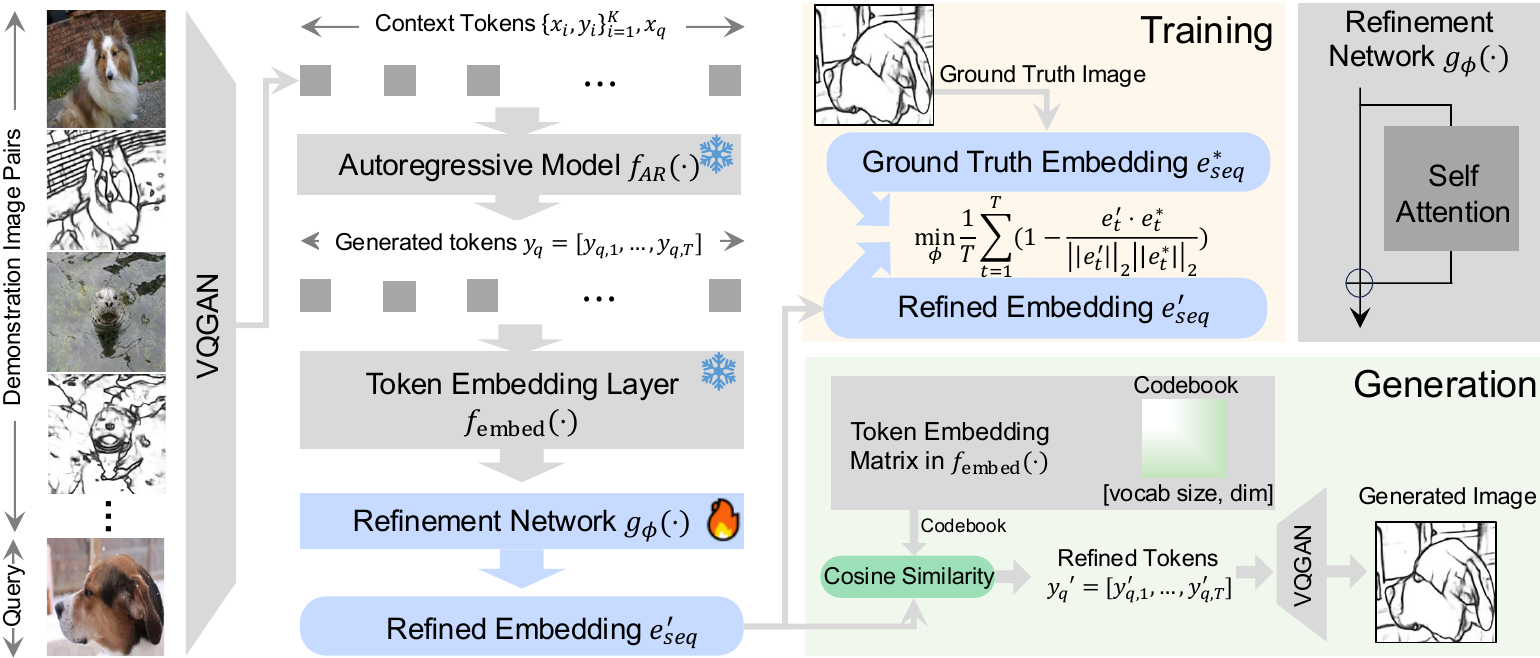}
  \vspace{-4mm}
  \caption{
  Training and generation pipeline of the proposed method. Input 
  images ($K$ demonstrations pairs and a query image) are encoded and fed into  $f_{AR}(\cdot)$ for generating tokens iteratively. A refinement network $g_\phi(\cdot)$ is introduced to improve the generated tokens $y_q$ in the embedding space, yielding $\mathbf{e}'_{seq}$. During \texttt{training}, we use ground truth embedding $\mathbf{e}^*_{seq}$ for supervision and minimize the cosine distance. During \texttt{generation}, the refined embedding $\mathbf{e}'_{seq}$ is decoded to the discrete tokens via nearest-neighbor search. Finally, the refined tokens $y'_q$ are passed to VQGAN decoder to generate the image. We take edge detection as an example. 
  }
  \label{fig:framework}
  \vspace{-3mm}
\end{figure*}

To address these challenges, we propose a plug-and-play self-refinement module that operates on the generated tokens while keeping the pretrained backbone frozen for the post-pretraining finetuning (see Figure~\ref{fig:framework}). Specifically, the proposed module jointly optimizes the generated tokens under the supervision of the target image. The discrete tokens are mapped to the embedding space to facilitate the learning. To this end, the broader view enables consistent corrections across the generated token sequence, mitigating errors stemming from the  prefix context in the autoregressive generation.

We summarize the contributions as follows: 
1) We introduce a post-pretraining finetuning operation to improve the next-token prediction for the visual data, without harming the generalizeability of the pretrained AR  model.
2) We propose to learn a plug-and-play lightweight module to facilitate the refinement,  with small-scale data and negligible running time cost.
3) Experiments demonstrate notable improvements and generalizeability of the proposed method  not only across various vision tasks but also on different pretrained AR backbones, with supporting empirical evidence showing improvement in token-wise prediction accuracy.

\section{Related Work}\label{sec: related work}

Autoregressive modeling serves as main streamline for image generation. LlamaGen~\cite{sun2024autoregressive} develops text-conditioned visual generation, AIM~\cite{el2024scalable} studies scalability in model capacity, data quantity and objective function, RAR~\cite{Yu2024RandomizedAV} rearranges token orders for bidirectional correspondence modeling, ELM~\cite{liu2024elucidating} studies the design space of LLMs for vision tasks, and VAR~\cite{tian2024visual} incorporates visual inductive bias via a next-scale prediction mechanism.

While these methods primarily focus on unconditional, and text/class-conditional generation, they fail to utilize images as context. LVM~\cite{bai2024sequential} address this gap by leveraging context image pairs for inferring tasks and generation, allowing a series of perception tasks~\cite{wang2023iterative} or reconstruction tasks~\cite{wang2022modeling,wang2025s,wang2024cooperative}. However, its reliance on next-token prediction leads to error accumulation and struggles to model spatial correspondences~\cite{tian2024visual}.

\section{Method}\label{sec: method}

\subsection{Preliminaries on LVM}\label{subsec: background}
\textbf{Autoregressive modeling.} aims to maximize the likelihood of a discrete token sequence $ \mathbf{x} = [x_1, x_2, \ldots, x_T] $ using a forward autoregressive factorization, 
where each token $ x_t $, \emph{i.e.}, discrete value index,  is predicted based 
on all preceding tokens $[x_1, x_2, \ldots, x_{t-1}]$:
\begin{equation}
    \max_\theta p_\theta(x) = \textstyle\prod_{t=1}^T p_\theta(x_t | x_1, x_2, \ldots, x_{t-1}),
\end{equation}
where $ p_\theta $ represents a token distribution predictor parameterized by $ \theta $. This sequential next-token prediction captures dependencies within the data and thus dynamically adapt to new tasks based on the contextual information.

\textbf{In-Context Visual Generation.}
LVM \cite{bai2024sequential} extended this next-token prediction modeling to different image translation and generation tasks through in-context learning. 
Given the input consisting 
 $ K $ demonstration pairs $ \{(\mathbf{x}_i, \mathbf{y}_i)\}_{i=1}^K $ and a query image tokens $ \mathbf{x}_{q} $ (e.g. rgb and sketch images  in Fig.~\ref{fig:framework}), the model dynamically adapts to the translation tasks such as inpainting or edge detection, and generates output tokens $\mathbf{y}_{q}=[y_{q,1}, y_{q,2}, \ldots, y_{q,T}]$ sequentially using a  autoregressive backbone $ f_{\text{AR}}$ with parameters $\theta$:

\begin{align}
y_{q,t} = 
\begin{cases} 
f_{\text{AR}}(\mathbf{x}_q, \{(\mathbf{x}_i, \mathbf{y}_i)\}_{i=1}^K; \theta), & t = 1, \\\\
f_{\text{AR}}(y_{q,<t}, \mathbf{x}_q, \{(\mathbf{x}_i, \mathbf{y}_i)\}_{i=1}^K; \theta), & t > 1,
\end{cases}
\end{align}
$ f_{\text{AR}} $ is iteratively applied to generate the target tokens $ y_{q} $, conditioned on the query and  context. 

\subsection{Next-Token Prediction for Visual Tokens}\label{subsec: limitations}
The next-token prediction generates tokens one by one in a fixed order, where each step  counts on the past context and  fits the natural languages with a sequential order. However, next-token prediction lacks a full view of the whole sequence at each step, being improper for the image data  whose information is spatially distributed and correlated~\cite{tian2024visual}. This makes it hard for the mechanism to maintain the global consistency and coherence for the generated image. Prior methods typically impose a predefined token generation order, such as raster-scan or spiral patterns~\cite{yu2024randomized}. However, these approaches generate each token based solely on preceding tokens and disregards useful information from future tokens that could provide additional contextual coherence. As a result, this sequential dependency leads to suboptimal outputs, particularly for complex visual tasks that requires holistic context of the entire image. 

\subsection{Self-Refinement for Visual Generation}\label{subsec: self_refinement}
To address these limitations, we propose a post-pretraining step to refine the generated visual token sequence $ \mathbf{y}_{q}$. Specifically, we introduce  a self-refinement module to jointly transform all the generated output tokens $\mathbf{y}_{q}=[y_{q,1}, y_{q,2}, \ldots, y_{q,T}]$ into refined tokens $\mathbf{y'}_{q}=[y'_{q,1}, y'_{q,2}, \ldots, y'_{q,T}]$, while keeping the autoregressive backbone frozen. By incorporating global context, the refinement module captures holistic relationships across all tokens, and thus significantly enhances the quality and coherence of the final visual output.
The refinement process involves: 1) refining the generated token embeddings with a self-attention module, 2) decoding the refined embeddings to discrete tokens via nearest-neighbor search.

\textbf{Self-Refinement Design.} 
The autoregressive generated tokens $\mathbf{y}_{q}$ are processed by pretrained embedding layer $ f_{\text{embed}}(\cdot) $ as the embeddings $\mathbf{e}_{seq} = [\mathbf{e}_1, \mathbf{e}_2, \ldots, \mathbf{e}_T]$. A refinement network $g_\phi$ parameterized by $\phi$, jointly processes these embeddings to produce refined embeddings $\mathbf{e'}_{seq} = [\mathbf{e'}_1, \mathbf{e'}_2, \ldots, \mathbf{e'}_T]$, which can be computed as:
\begin{align}
    \mathbf{e'}_{seq} = \mathbf{e}_{seq}+ \mathrm{SelfAttention}(\mathbf{e}_{seq}; \phi)
\end{align}
The self-attention module captures spatial relationships, semantic coherence and long-range dependencies across the sequence. Importantly, processing all the embeddings simultaneously make the refinement step to adjust for inconsistencies and deviations introduced during sequential generation. 

\textbf{Self-Refinement Optimization.}
For post-pretraining fine-tuning, the refinement module $g_\phi(\cdot)$ is optimized and operates as an plug-and-play component upon the pretrained LVM. Specifically, we jointly consider all the generated tokens in the latent space and  minimize the cosine distance between the sequence of refined embeddings $\mathbf{e'}_{seq}$ and the ground truth token embeddings $\mathbf{e}^*_{seq}$:
\begin{align}
    \min_\phi \frac{1}{T} \sum_{t=1}^T \left( 1 - \frac{\mathbf{e'}_t \cdot \mathbf{e}_t^*}{\|\mathbf{e'}_t\|_2 \|\mathbf{e}_t^*\|_2} \right),
\end{align}
which provides the supervision for the generated embedding to align with the target distribution, improving the fidelity and coherence of the output with small-scaled data.

\textbf{Generation.} 
A decoding step is then performed to transform the refined embeddings $\mathbf{e'}_{seq}$ into discrete tokens $\mathbf{y'}_{q}$ to ultimately reconstruct the output image. Given the embeddings of all tokens in the codebook (\emph{i.e.}, embedding matrix from pretrained $f_\text{embed}(\cdot)$), we identify the nearest token embedding to each refined embedding based on cosine similarity. The token associated with the closest token embedding is selected as the refined token. Once determined, the refined tokens are decoded into images using the VQGAN decoder.

\section{Experiment}\label{sec: experiment}
We evaluate our method on three vision tasks: image colorization, inpainting, and edge detection.

\textbf{Datasets.} 
We adopt the large-scale Unified Vision Dataset (UVDv1) curated by LVM~\cite{bai2024sequential} for each task. The proposed self-refinement module is tuned with small-scale data (\emph{e.g.}, $12$K image pairs). The image resolution is $256\times256$.

\textbf{Baselines.} 
We compare with three prevalent baselines. We adopt pretrained LVM (LLaMA-7B) \citep{bai2024sequential} for vanilla in-context generation (\texttt{LVM}). We include one baseline (\texttt{+Context Retrieval}) that enhances the generation with context images retrieved based on similarity-search \citep{zhang2023makes}. We include another baseline (\texttt{+LoRA}) that finetunes LVM with LoRA~\cite{hu2021lora} on each task. We apply LoRA with the rank of 8 and insert them into query and value projection layers of the multi-head attention modules in each Transformer block.  Our  method only learns the refinement network $g_\phi(\cdot)$ without LoRA module (\texttt{+Self-Refinement}). Both LoRA fine-tuning and our method are trained with  batch size of $4$ using AdamW optimizer and learning rate $1e^{-4}$ for $2$ epochs. We adopt $4$ context pairs for all tasks. We provide more details on metrics in the appendix.

\begin{table}[t]
\centering
\renewcommand{\arraystretch}{1.2} 
\caption{Comparison of our self-refinement method against baselines across different vision tasks.}
\label{tab:stacked_results}
\vspace{-2mm}

\begin{subtable}[t]{\linewidth}
    \centering
    \tiny
    \begin{tabularx}{\linewidth}{l|*{6}{>{\centering\arraybackslash}X}}
    \hline
    \multirow{2}{*}{\textbf{Method}} & \multicolumn{6}{c}{\textbf{Colorization}} \\
    \cline{2-7}
    & \textbf{Perp} $\downarrow$ & \textbf{LPIPS} $\downarrow$ & \textbf{FID} $\downarrow$ & \textbf{IS} $\uparrow$ & \textbf{PSNR} $\uparrow$ & \textbf{SSIM} $\uparrow$ \\
    \hline
    LVM              & 20.06 & 0.29 & 59.70 & 27.20 & 17.93 & 0.4925 \\
    +Context Retrieval & 19.94 & 0.30 & 59.63 & 26.59 & 17.87 & 0.4936 \\
    +LoRA            & 19.04 & 0.29 & 59.76 & 27.14 & 18.09 & 0.5005 \\
    \rowcolor{gray!18}
    +Self-Refinement & \textbf{19.01} & \textbf{0.28} & \textbf{59.24} & \textbf{27.34} & \textbf{18.78} & \textbf{0.5086} \\
    \hline
    \end{tabularx}
\end{subtable}

\vspace{1.5mm} 

\begin{subtable}[t]{\linewidth}
    \centering
    \tiny
    \begin{tabularx}{\linewidth}{l|*{6}{>{\centering\arraybackslash}X}}
    \hline
    \multirow{2}{*}{\textbf{Method}} & \multicolumn{6}{c}{\textbf{Inpainting}} \\
    \cline{2-7}
    & \textbf{Perp} $\downarrow$ & \textbf{LPIPS} $\downarrow$ & \textbf{FID} $\downarrow$ & \textbf{IS} $\uparrow$ & \textbf{PSNR} $\uparrow$ & \textbf{SSIM} $\uparrow$ \\
    \hline
    LVM              & 175.93 & 0.31 & 63.95 & 25.35 & 17.38 & 0.4340 \\
    +Context Retrieval              & 183.22 & 0.31 & 64.31 & 23.79 & 17.41 & 0.4356 \\
    +LoRA       & 172.80 & 0.31 & 64.42 & 25.49 & 17.54 & 0.4395 \\
    \rowcolor{gray!18}
    +Self-Refinement & \textbf{82.63} & \textbf{0.27} & \textbf{59.60} & \textbf{26.71} & \textbf{18.40} & \textbf{0.4508} \\
    \hline
    \end{tabularx}
\end{subtable}

\vspace{1.5mm} 

\begin{subtable}[t]{\linewidth}
    \centering
    \scriptsize
    \begin{tabularx}{\linewidth}{l|>{\centering\arraybackslash}X>{\centering\arraybackslash}X>{\centering\arraybackslash}X}
    \hline
    \multirow{2}{*}{\textbf{Method}} & \multicolumn{3}{c}{\textbf{Edge Detection}} \\
    \cline{2-4}
    & \textbf{Perp} $\downarrow$ & \textbf{Acc} $\uparrow$ & \textbf{Recall} $\uparrow$ \\
    \hline
    LVM              & 23.53 & 0.76 & 0.80 \\
    +Context Retrieval & 23.24 & 0.76 & 0.81 \\ 
    +LoRA         & 22.17 & \textbf{0.84} & 0.86 \\
    \rowcolor{gray!18}
    +Self-Refinement & \textbf{20.00} & 0.83 & \textbf{0.88} \\
    \hline
    \end{tabularx}
\end{subtable}
\vspace{-2mm}
\end{table}

\begin{table}[t]
\centering
\caption{
Self-refinement for the image generation. We fix the VAR and only fine-tune self-refinement module.}
\vspace{-2mm}
\scalebox{0.78}{ 
\begin{tabular}{lcc}
\hline
{Metrics} & {Vanilla VAR} & {VAR w/ self-refinement (ours)} \\
\hline
FID $\downarrow$ & 3.55 & \textbf{3.43} \\
IS $\uparrow$ & 284.2 & \textbf{287.4} \\
Precision $\uparrow$ & 0.85 & \textbf{0.87} \\
Recall $\uparrow$ & 0.49 & \textbf{0.51} \\
\hline
\end{tabular}
}
\vspace{-6mm}
\label{tab:self_refinement}
\end{table}

\begin{table}[t]
\centering
\caption{
Performance comparison under structural-error scenario on colorization.}
\vspace{-1mm}
\scalebox{0.7}{ 
\begin{tabular}{lcccccc}
\hline
Methods & Perplexity & LPIPS & FID & IS & PSNR  \\
\hline
w/o Self-refinment &	23.70 & 0.35 & 65.85 & 25.00 & 16.07  \\
w/ Self-refinment  & 20.53 & 0.30 &	60.10 &	26.90 &	17.82 \\
\hline
\end{tabular}
}
\label{tab:structured_error}
\end{table}

\subsection{Performance Comparison}\label{subsec: performance}
\textbf{Image Perception}.  In Table~\ref{tab:stacked_results}, we demonstrate the results of our self-refinement method and baselines on colorization, inpainting and edge detection tasks. Our method demonstrates superior performance across all three tasks.
For colorization, our method achieves the best performance across all metrics, demonstrating superior visual quality and consistency. For inpainting, our self-refinement method significantly improves perplexity (82.63), far outperforming all other methods. It also delivers the best LPIPS, FID, IS, PSNR, and SSIM, demonstrating its robustness and superiority in reconstructing realistic and coherent content. On edge detection, our method achieves the lowest perplexity, indicating superior model confidence, and the highest recall, showcasing its ability to capture more relevant edges. While its accuracy (0.83) is slightly below LoRA finetuning (0.84), it maintains better perplexity and recall, demonstrating a balanced and robust performance. Note that the number of the trainable parameters is $14.7$M for LoRA modules and $16.1$M for the proposed self-refinement module, being comparable regarding the model size. 

\textbf{Image Generation}. 
In Table~\ref{tab:self_refinement}, we extend the proposed self-refinement design to the visual autoregressive model of VAR~\cite{tian2024visual} under the task of class-conditional generation.  We fine tune the  self-refinement module on ImageNet 256x256 and evaluation on its benchmark dataset. Overall, the proposed method demonstrates the generalizability on (1) different AR models and (2) on different tasks including perception and generation. We set the same temperature and the seed as the vanilla VAR to improve the statistical stability for the generation.

\begin{figure}[t]
  \includegraphics[width=\columnwidth]{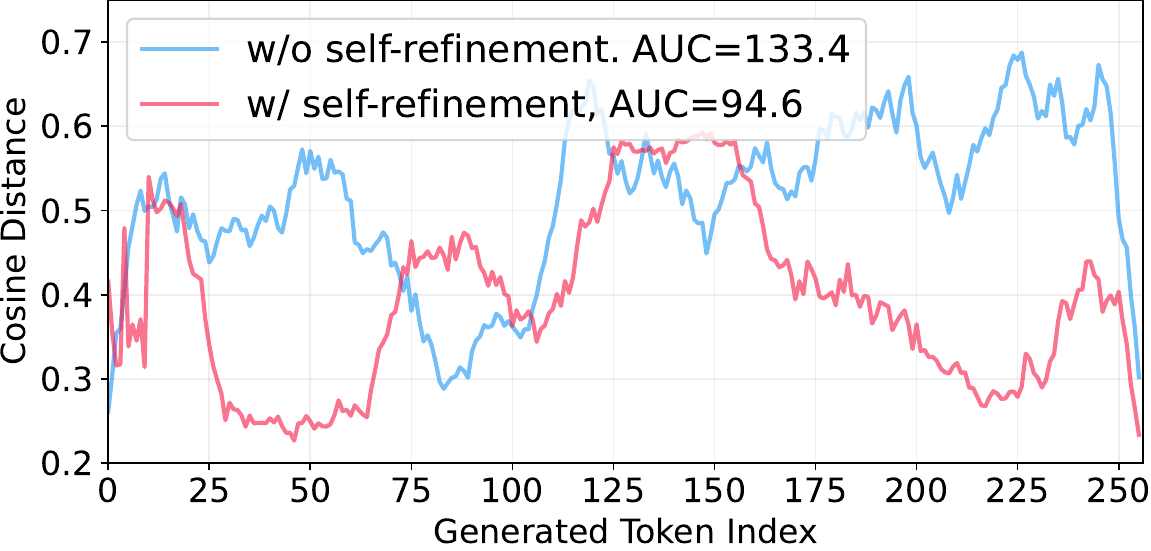}
  \vspace{-6.5mm}
  \caption{
  We compare the LVM with LoRA and the LVM with self-refinement in terms of the cosine distance among generated embeddings and the ground truth embeddings for $256$ tokens. Self-refinement reduces the error accumulation especially for latter tokens ($>160$). The accumulated errors (\emph{i.e.}, by AUC) are provided.  
  }
  \label{fig:loss_comparison}
  \vspace{-3mm}
\end{figure}

\textbf{Error Accumulation Discussion.}
Fig.~\ref{fig:loss_comparison} compares the cosine distance of generated embeddings and the ground-truth embeddings between the LVM with LoRA and with self-refinement under inpainting. LVM with self-refinement achieves significantly reduced cosine distance for early tokens ($25<t<70$) and later tokens ($t>160$), enabling smaller AUC value. The results highlight that the self-refinement process corrects early errors and mitigates error accumulation during generation. For the quantitative comparison, we feed the first 70 refined tokens (which manifest reduced cosine distance) into the LVM to generate the rest tokens, and then comparing the quality of the remaining generated tokens with and without refinement in Table~\ref{tab:remain_token}. Both the token-level evaluation (e.g., accuracy for the remaining generated tokens) and the image-level evaluations are improved with the proposed method. The self-refinement module works differently with the next-token prediction by jointly optimizing the entire generated sequence and adjusting the structural inconsistency globally. Thus better handle the potential fundamental mistakes,  improving the quality of the discrete tokens and alleviates the error accumulation.

\textbf{Structured Error Discussion.}
The proposed module works differently with the next-token prediction by jointly optimizing the entire generated sequence and adjusts the structural inconsistency globally. Thus better handle the potential fundamental mistakes that the next-token prediction could not solve. In Table~\ref{tab:structured_error}, we simulate the scenario with structural error by locally permuting a segment of the token embeddings (randomly selecting $10\%$ of the continuous tokens from the generated sequence) and compare the performance after decoding. The vanilla LVM (``w/o self-refinement’’) experiences large performance descent compared to the baseline in Table~\ref{tab:stacked_results}, indicating severe sequential prediction errors. The proposed self-refinement module brings remarkable performance boost.

\textbf{Breakdown Running Time Analysis}. 
We compute the running time on a workstation with 2x NVIDIA RTX A6000 GPUs, Intel Core i9-10900X @ 3.70GHz, 10 cores / 20 threads CPU and system memory of 128GB.
As shown in Table~\ref{tab:time_breakdown}, the self-refinement module is lightweight and only requires $0.112$s to process. We perform nearest-neighbor lookup for $256$ generated tokens in parallel, thus being efficient. The token generation induces most time cost and depends on the number of tokens in the input sequence, which does not affect the self-refinement.

\begin{table}[t]
\centering
\caption{
Quality comparison of the remaining generated tokens on colorization.}
\vspace{-2mm}
\scalebox{0.78}{ 
\begin{tabular}{lcc}
\hline
{Metrics} & {w/o Self-refinment} & {w/ Self-refinement} \\
\hline
Gen Token Acc $\uparrow$ & 0.667 & \textbf{0.694} \\
Perplexity $\downarrow$ & 20.06 & \textbf{19.50} \\ 
LPIPS $\downarrow$ & 0.29 & 0.29 \\
FID $\downarrow$ & 59.70 & \textbf{59.66} \\
IS $\uparrow$ & 27.20 & \textbf{27.28} \\
PSNR $\uparrow$ & 17.93 & \textbf{18.53} \\
SSIM $\uparrow$ & 0.4925 & \textbf{0.5054} \\
\hline
\end{tabular}
}
\vspace{-2mm}
\label{tab:remain_token}
\end{table}

\begin{table}[t]
\centering
\caption{
Breakdown running time (s) analysis under $k=4$ demonstration pairs. We compute average generation time (per image) and compare under the same  platform.}
\vspace{-2mm}
\scalebox{0.75}{
\begin{tabular}{lcc}
\hline
\textbf{Operations} & \textbf{LVM Baseline} & \textbf{Ours} \\
\hline
Token generation & 4.578 & 4.578 \\
Self-Refinement module processing & -- & 0.112 \\
Nearest-neighbor look-up & -- & 0.155 \\
VQGAN decoding & 0.233 & 0.233 \\
\hline
\textbf{Total time cost (s)} & \textbf{4.811} & \textbf{5.078} \\
\hline
\end{tabular}
}
\vspace{-4.5mm}
\label{tab:time_breakdown}
\end{table}

\section{Conclusion}\label{sec: conclusion}

This work proposed a lightweight post-processing technique to improve the next-token generation for the image perception and generation. By jointly refining the tokens, the proposed method jointly incorporated the global contextual information in the image, mitigating the error accumulation in vanilla next-token prediction without affecting the generative capabilities of the pretrained AR model, enhancing the visual coherence and the performance.

\paragraph{Limitations.} 
As a post-processing step, refinement process does not directly influence the initial autoregressive generation. Errors introduced during the sequential token generation stage are mitigated with refinement but not eliminated. Future works could explore tighter integration between refinement and generation for improved results. Besides, the generalization and portability to other problems could be validated.

\bibliography{main}

\clearpage
\appendix
\section{Appendix}
\label{sec:appendix}

\begin{figure*}[t] 
\centering
\begin{minipage}{0.95\textwidth}
\begin{algorithm}[H]
\caption{Training with Refinement Network}
\begin{algorithmic}[1]
\STATE \textbf{Input:} Context image pairs $\mathbf{x}_1, \mathbf{y}_1, \dots, \mathbf{x}_K, \mathbf{y}_K$, query image $\mathbf{x}_q$, ground truth label image $\mathbf{x}_\text{GT}$
\STATE \textbf{Modules:} VQGAN tokenizer $T$, AR model $f_{AR}(\cdot)$, refinement network $g_\phi(\cdot)$, loss function $\mathcal{L}$
\STATE \textbf{Output:} Refined token embeddings $\mathbf{e}'_{\text{seq}}$
\STATE Encode all images into token sequences: $\textit{Input Ids} \leftarrow T.\text{encode}([\mathbf{x}_1, \mathbf{y}_1, \dots, \mathbf{x}_K, \mathbf{y}_K, \mathbf{x}_q])$
\STATE Generate output tokens with AR model: $\mathbf{y}_q \leftarrow f_{AR}(\textit{Input Ids})$
\STATE Embed predicted tokens: $\mathbf{e} \leftarrow f_{\text{embed}}(\mathbf{y}_q)$
\STATE Refine embeddings: $\mathbf{e}' \leftarrow g_\phi(\mathbf{e})$
\STATE Compute target embeddings: $\mathbf{e}^* \leftarrow f_{\text{embed}}(T.\text{encode}(\mathbf{x}_\text{GT}))$
\STATE Compute loss: $\mathcal{L} \leftarrow \text{CosineDistance}(\mathbf{e}', \mathbf{e}^*)$
\STATE Backpropagate to update parameters of $g_\phi$ only
\end{algorithmic}
\end{algorithm}
\end{minipage}
\end{figure*}


\begin{table*}[h]
\centering
\small
\caption{Distance measuring discussion for self-refinement network learning, \emph{e.g.}, upon colorization and inpainting.}
\begin{tabularx}{\textwidth}{c|*{6}{>{\centering\arraybackslash}X}|*{6}{>{\centering\arraybackslash}X}}
\hline
\multirow{2}{*}{Distance} & \multicolumn{6}{c|}{\textbf{Colorization}} & \multicolumn{6}{c}{\textbf{Inpainting}} \\
\cline{2-13}
& \textbf{Perp} & \textbf{LPIPS} & \textbf{FID} & \textbf{IS} & \textbf{PSNR} & \textbf{SSIM} 
& \textbf{Perp} & \textbf{LPIPS} & \textbf{FID} & \textbf{IS} & \textbf{PSNR} & \textbf{SSIM} \\
\hline
{Cosine} & 19.01 & 0.28 & 59.24 & 27.34 & 18.78 & 0.5086 & 82.63 & 0.27 & 59.60 & 26.71 & 18.40 & 0.4508 \\
{$L_2$} & 20.06 & 0.30 & 59.75 & 27.24 & 17.89 & 0.4922 & 176.00 & 0.31 & 64.05 & 25.40 & 17.24 & 0.4337 \\
\hline
\end{tabularx}
\label{tab:discussion_dist}
\end{table*}

\begin{table*}[h]
\centering
\scalebox{0.8}{
\begin{tabular}{lcccc}
\toprule
\textbf{Metric} & \textbf{LVM} & \textbf{+MLP} & \textbf{+Convolutional network} & \textbf{+Self-Attention (proposed)} \\
\midrule
Perplexity $\downarrow$     & 20.06   & 19.70   & 19.26   & \textbf{19.01} \\
LPIPS $\downarrow$    & 0.29    & 0.30    & 0.29    & \textbf{0.28} \\
FID $\downarrow$      & 59.70   & 60.95   & 59.40   & \textbf{59.24} \\
IS $\uparrow$         & 27.20   & 26.50   & 27.00   & \textbf{27.34} \\
PSNR $\uparrow$       & 17.93   & 17.90   & 18.30   & \textbf{18.78} \\
SSIM $\uparrow$       & 0.4925  & 0.4800  & 0.5001  & \textbf{0.5086} \\
\bottomrule
\end{tabular}
}
\caption{
Ablation study of refinement network structure on colorization.}
\label{tab:refinement_ablation}
\end{table*}

\begin{figure*}[t]
    \centering
    \includegraphics[width=0.95\linewidth]{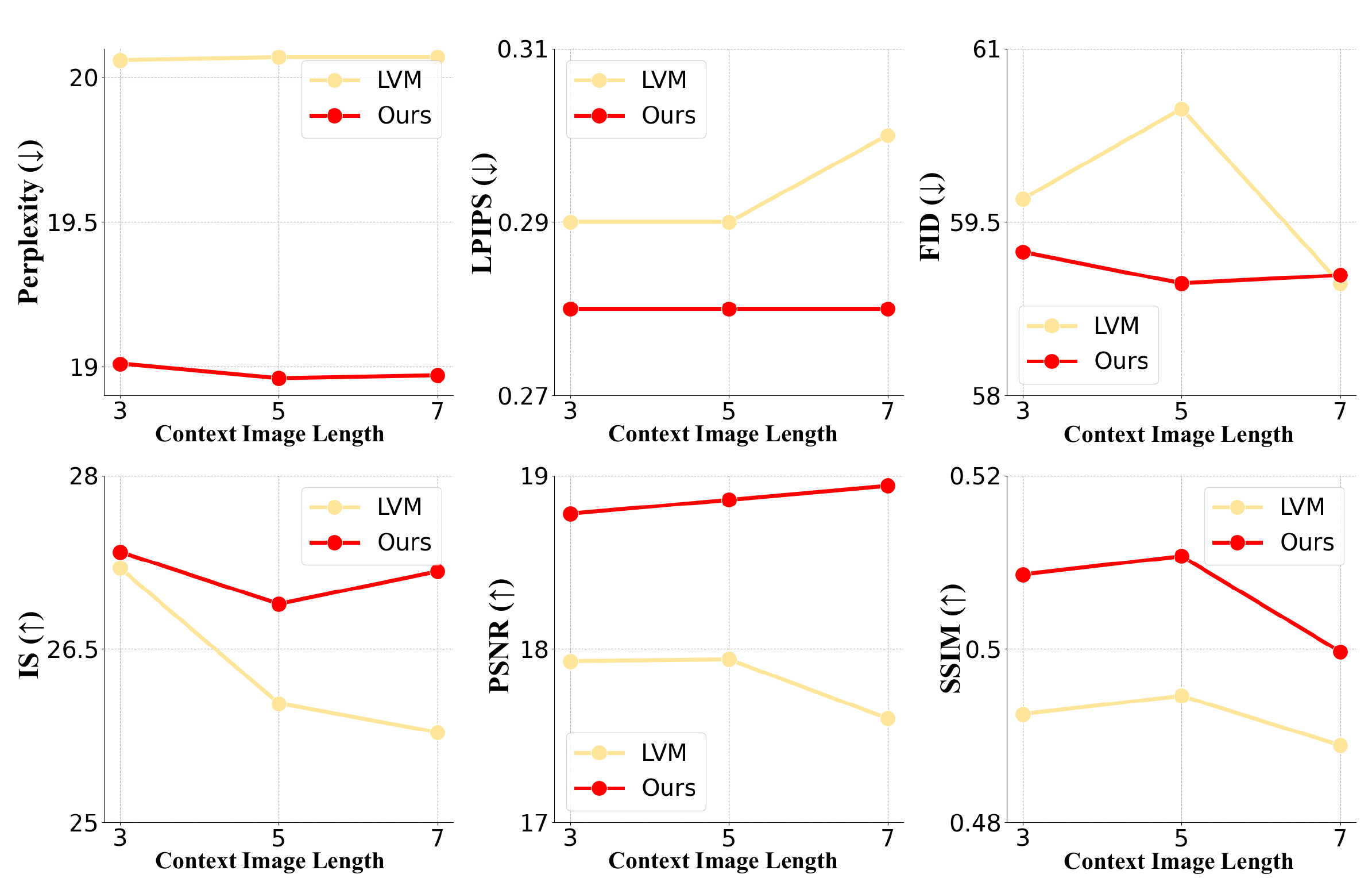}
    \caption{Effect of demonstration image length for the proposed self-refinement (\emph{e.g.},  upon colorization). The input image sequence is composed of one or more image pairs plus one query RGB image. The proposed method enables notable performance boost under different context lengths, demonstrating the robustness to the context length.}
    \label{fig:context_length}
\end{figure*}

\textbf{Metrics.} 
On colorization and inpainitng, we employ the Fréchet inception distance (FID)~\cite{heusel2017gans}, Inception Score (IS)~\cite{salimans2016improved}, Perplexity, and LPIPS~\cite{zhang2018unreasonable} to measure the generation quality, and incorporate  PSNR~\cite{korhonen2012peak} (dB) and SSIM~\cite{1284395} to assess the image quality comparing to ground-truth image. On edge detection, we measure the performance in terms of detection accuracy and recall.

\textbf{UVDv1 Dataset}.
As introduced in Section~\ref{sec: experiment}, we adopt the UVDv1 dataset curated by LVM~\cite{bai2024sequential}, which covers images from over 50 large-scale datasets including ImageNet, COCO, and LAION, etc. and  inherits the multi-source nature. We split $12$K image pairs for training and $1$K for inference in each task. Notably, the dataset avoids randomness and ensures stability for two reasons. (1) We adopt a fixed split of image pairs for training across all experiments, ensuring the consistency and stability. (2) UVDv1 dataset covers a wide range of visual variations and ensuring that the initial selection in the split does not affect the reliability of the training.

\textbf{More Details on LVM}. 
LVM~\cite{bai2024sequential} fine tunes the VQGAN tokenzier for image encoding and reconstruction.  For visual in-context learning, LVM generates the image token sequence $\mathbf{y}_q$ for the given query image token sequence $\mathbf{x}_q$ depending on  tasks defined by the given $K$ demonstration pairs, \emph{i.e.}, $\{\mathbf{x}_i, \mathbf{y}_i\}^K_{i=1}$, where each pair consiste of a RGB image token sequence $\mathbf{x}_i$ and a sketch image token sequence $\mathbf{y}_i$ (exampled by edge detection). For each token sequence (e.g., $\{\mathbf{x}_i, \mathbf{y}_i, \mathbf{x}_q, \mathbf{y}_q\}$), there are $256$ tokens set by LVM. Owning to this property, LVM readily enables diverse image translation tasks or generation tasks.

\textbf{More Details on VQGAN Tokenizer}. 
In VQGAN tokenizer~\cite{esser2021taming}, each image is divided into non-overlapping patches (e.g., for LVM), and each patch is encoded into a latent vector by the pretrained VQGAN encoder. These vectors are then quantized by mapping to a discrete token. Specifically, A token refers to a discrete index value.  These tokens are good representations as (1) they can reflect meaningful visual patterns like edges, textures, or object parts by capturing the local visual structures. (2) Their discrete nature allows autoregressive modeling, facilitating the image generation.

\textbf{More Details on Decoding}. 
In the baseline method of LVM, each image will be encoded into $256$ discrete tokens. Besides, the ``codebook'' refers to the embedding matrix provided by the token embedding layer $f_\text{embed}(\cdot)$
, which defines the mapping between discrete tokens and their corresponding embedding vectors. The embedding layer comes from the pretrained LVM model and is kept fixed throughout the training, as indicated by the ``frozen’’ symbol in Fig.~\ref{fig:framework}. It is also used during generation as shown in Fig.~\ref{fig:framework} (lower right block). Based on the codebook, the continuous refined embedding is mapped back to the discrete tokens via nearest-neighbor lookup.

\textbf{More Technical Details of the Self-Refinement}. The proposed method jointly adjusts the discrete tokens in the latent space. To achieve this, we firstly map the generated distrete tokens to the embedding space for fine-tuning and map the embedding back to the discrete token space for generation (inference). Specifically, as shown in Fig.~\ref{fig:framework}, during training, the loss is computed purely under the embedding space. No nearest embedding lookup is needed. Differently, during inference, it involves finding the nearest embedding in the codebook to reconstruct the image.

The proposed self-refinement network $g_\phi(\cdot)$ adopts a Transformer architecture with multi-head self-attention and residual connections. Each block consists of (1) MultiHead-Attention + Add \& LayerNorm and (2) Feedforward Network (FFN) + Add \& LayerNorm. The full refinement network stacks multiple such blocks (e.g., \#head$=8$, \#blocks$=1$). This structure enables the model to capture global dependencies across generated token embeddings and the joint refinement. In Table~\ref{tab:refinement_ablation}, 
we discuss different network structures of the proposed self-refinement module, including self-attention, MLP, and 1D convolutional structure. All variants preserve residual structure and comparable parameter scale. Self-attention significantly outperforms both MLP and CNN. MLP does not bring the performance boost for LVM baseline, due to its token-wise nature. Convolutional network enables local structure modeling and improve over MLP but remain less effective than self-attention.

\textbf{More Details on Image Generation}. 
We choose VAR-d16 as the baseline model. The experiments are conducted on 2 A6000 GPUs. Table~\ref{tab:self_refinement} shows that the proposed self refinement module enables further performance boost compared to pretrained VAR after training for 10K iterations.

\textbf{Context Length Discussion}.  In Figure~\ref{fig:context_length}, we provide a detailed comparison of our self-refinement method (red) against the ICL baseline (yellow) across various metrics on colorization task as the context image length increases. The results demonstrate the consistent superiority of our approach. As shown in first row in Figure~\ref{fig:context_length}, our method consistently achieves significantly lower perplexity, LPIPS, and FID across nearly all context image lengths, demonstrating more confident predictions and better fidelity as well as alignment with real-world image distributions. While for IS, PSNR and SSIM, our self-refinement method maintains higher metric score compared with ICL baseline. We also observe that increasing the context image length does not always lead to improved performance, a trend that aligns with similar findings in LVM~\cite{bai2024sequential}. 

\textbf{More Ablation Studies}.  In Table~\ref{tab:discussion_dist}, we compare different similarity metrics to be used for learning the refinement network $g_\phi(\cdot)$. As shown by both colorization and inpainting, cosine distance enables promising performance and can be better than $L_2$ distance. This is because during decoding, VQGAN model adopts a cosine similarity as the distance measuring to identify the closest element in codebook for the given input. Accordingly, bridging the embedding with the cosine distance benefits the token identification. 

\textbf{Social Impacts}. Like other generative models, the proposed method could be misused to create misleading or harmful content if applied irresponsibly. However, our focus is on improving controllability and quality in image generation, which can benefit creative and commercial applications. 

\end{document}